# Predicting Road Crossing Behaviour using Pose Detection and Sequence Modelling

Subhasis Dasgupta
subhasisdasgupta1@acm.org
Praxis Business School,
Kolkata

Preetam Saha
preetam.saha@praxis.ac.in
Praxis Business School,
Kolkata

Agniva Roy
agniva.roy@praxis.ac.in
Praxis Business School,
Kolkata

Jaydip Sen
jaydip.sen@acm.org
Praxis Business School,
Kolkata

*Abstract: The world is constantly moving towards AI based systems and autonomous vehicles are now reality in different parts of the world. These vehicles require sensors and cameras to detect objects and manoeuvre according to that. It becomes important to for such vehicles to also predict from a distant if a person is about to cross a road or not. The current study focused on predicting the intent of crossing the road by pedestrians in an experimental setup. The study involved working with deep learning models to predict poses and sequence modelling for temporal predictions. The study analysed three different sequence modelling to understand the prediction behaviour and it was found out that GRU was better in predicting the intent compared to LSTM model but 1D CNN was the best model in terms of speed. The study involved video analysis, and the output of pose detection model was integrated later on to sequence modelling techniques for an end-to-end deep learning framework for predicting road crossing intents.*

## I. Introduction

The world is rapidly advancing toward a future where artificial intelligence (AI) takes a central role in many everyday activities. In business, for example, robots have become indispensable in manufacturing processes and warehouse management. These robots efficiently handle tasks such as stacking and removing items, optimizing various business operations. In aviation, autopilot systems have been a standard feature in airplanes for many years, enhancing flight safety and efficiency. Similarly, in many developed countries, vehicles equipped with autopilot capabilities are becoming increasingly common. These self-driving vehicles are designed with an array of sensors and high-resolution cameras to monitor their surroundings, detect objects, and take necessary actions to prevent collisions or accidents.

While these autonomous vehicles perform admirably on highways where the primary concern is other vehicles, they face significant challenges in busy urban environments. In such settings, it is often advisable for drivers to switch from autopilot to manual control. This is particularly crucial in bustling market areas where pedestrian behaviour can be unpredictable. People crossing the road in these scenarios often exhibit erratic and seemingly irrational behaviours. If a self-driving car is not adequately equipped to anticipate these actions, the risk of accidents increases, potentially leading to severe injuries or loss of life. The current project aims to address this critical issue by developing a system to predict pedestrian intent to cross the road based on their movements. The goal is to enhance the decision-making capabilities of autonomous vehicles, particularly in complex urban environments. To achieve this, an experimental setup was created where participants were asked to simulate road-crossing behaviours. These behaviours were recorded on video for further analysis.

In the experimental setup, participants' movements were meticulously recorded to capture various nuances of road-crossing behaviour. These recordings provided a rich dataset that allowed the researchers to train their models to recognize subtle cues and patterns indicative of a person's intent to cross the road. By analysing these sequences of movements, the models learned to differentiate between genuine crossing attempts and other movements that might not result in crossing the road.

The deep learning models used in this project leveraged state-of-the-art techniques to process and interpret the video data. This included using convolutional neural networks (CNNs) to extract spatial features from the frames and recurrent neural networks (RNNs) to understand the temporal dynamics of the movements. Long Short-Term Memory (LSTM) networks and Gated Recurrent Units (GRUs) were particularly effective in this regard due to their ability to maintain and update a memory of previous frames, helping to predict future actions accurately.

The current study shows the usage of pose detection methods along with sequential modelling to understand the intent of road crossing behaviour. There are prior works available to predict the same behaviour at zebra crossing but not at open roads where pedestrians at times take risks to cross the road amid vehicle movements. The present study tries to mimic the prediction made by a human driver looking at the pedestrian even when he/she is having movements at the edge of the road.

In that sense, this work is different from the available works in this domain. The subsequent parts of the paper are organized as follows. In section II, some details are given about the relevant past works and their mode of operations. Section III explains the methodology in details. Section IV is the actual analysis part with results of the experiments. Finally, section V gives the conclusion remarks.

## II. LITERATURE REVIEW

Predicting pedestrian movement is important from the point of view of managing traffic, particularly, at the crossings. Even though roads have got traffic signals, yet not everywhere such signals are found. Particularly, in developing countries, like India, even though there are good amount of coverage but that is not enough. Keeping this in mind, different researchers have proposed their studies earlier to predict this behaviour. A critical approach to predict this behaviour is to understand the poses of the pedestrians and analyse them to determine if the person is crossing the road or not. Camara et al. tried to predict the assertiveness of the pedestrian to cross the road by using probabilistic model like logistic regression and decision tree model [1]. Kotseruba et al. did a study to understand the same behaviour using pedestrian trajectory [2]. Zhang and Yuan [3] tried to predict the road crossing behaviour based on object tracking and LSTM [4] model. Pose detection is another area in the domain of computer vision and it has been used in several different research [5-7]. Zhao et al. propose a novel method combining data from 3D LIDAR and cameras to enhance object detection capabilities in autonomous vehicles [8]. By integrating the precise depth information from LIDAR with rich color and texture data from cameras, the scheme improves the accuracy and reliability of detecting and identifying objects in various driving environments. Su et al. introduce the Spatial Interaction Transformer (SIT), a generative model that captures spatio-temporal correlations of pedestrian movements using a transformer-based architecture [9]. SIT effectively models interactions among pedestrians, leading to improved accuracy in trajectory prediction tasks. Alahi et al. present a novel LSTM model to predict human trajectories by capturing social interactions in crowded environments [10]. The model incorporates a unique pooling layer that aggregates hidden states from neighbouring LSTMs. This enables the system to account for the influence of nearby individual on one's movement. Zong et al. propose a model that integrates social interactions and environmental context to predict pedestrian movements [11]. The scheme utilizes a graph attention network to capture dynamic relationships among pedestrians and incorporates scene information to account for obstacles and preferred pathways. Vizzari & Cecconello introduce a method that employs reinforcement learning to model pedestrian behavior [12]. Utilizing a curriculum-based training strategy within a Unity and ML-Agents framework, the approach proposed by the authors incrementally exposes agents to complex scenarios. This enhances the adaptability of the agents to various situations. The resulting pedestrian behavioural model demonstrates generalization capabilities and hence, it effectively handles scenarios which were not encountered during training. Gazzeh & Douik demonstrate a real-time system designed to recognize pedestrians' intentions to cross the street [13]. The approach leverages deep learning techniques to analyze pedestrian behaviour, thereby enabling vehicles to anticipate crossing actions and respond accordingly. For the current study, it was required that a solution be chosen in such a way that the existing pretrained models stayed updated. Keeping this in mind, MediaPipe was chosen as one of the python packages for pose detection as this package is maintained by Google AI team. There are research involving the use of MediaPipe such as sign language detection [14], hand tracking [15] and suspicious human movement detection [16].

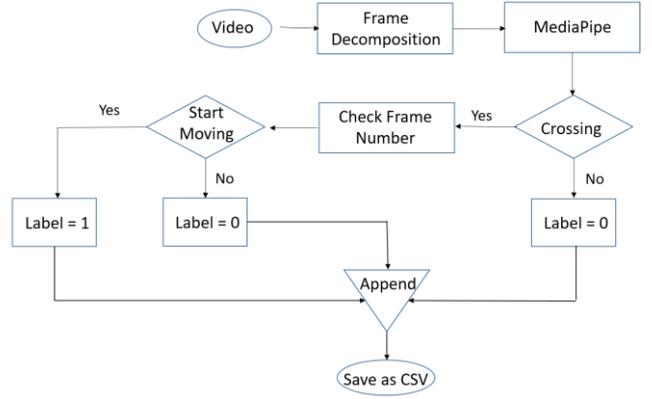

**Figure 1**: Training data preparation flowchart

For the current study, the objective was similar to that of [3] but instead of tracking the trajectory, MediaPipe was considered to work with only landmark points to understand the intent of the pedestrian to cross the road based on the movements of the landmark points. Intent prediction was considered important for the study so that corrective measures can be taken if the intent is strong enough (probability of crossing > 50%) to cross the road. In the current study, MediaPipe was used in combination with LSTM, GRU and 1D CNN to obtain the desired result.

## III. METHODOLOGY

Predicting whether a person is going to cross a road or not is not that simple as it is nearly impossible to understand what is going on in the mind of the person before he makes a move. However, common movements can provide important cues to probabilistically predict the intent of the pedestrian to cross the road. However, in this study, the scope was rather restricted to a single person who may or may not try to cross the road. This was done to see if it worked on a single pedestrian because on successful implementation, it could be scaled to adopt multiple pedestrians. It was assumed that human poses could give an indication whether the person was about to cross the road or not. Hence, pose detection played a critical role in this study. The methodology is further broken down into two subsections as follows:

*Data Collection:* The current study needed data in sequential manner. The data needed for building the model were pose anchor points of the pedestrian. Initially, for this purpose, the PIE dataset [17] was considered containing large volume of videos consuming almost 74 GB of memory. Dealing with such large volume of data was not possible due to resource constraint. Hence, for this study, primary data were generated by capturing videos with handheld mobile phones. Around 60 video clips were collected where a single person was trying to cross the road from either right to left or from left to right. 20 different people volunteered to provide the short videos where they mimicked the road crossing tendencies with different postures. Moreover, there were videos when the pedestrian was making movements but not crossing the road. To maintain the consistency of resolution, only a single camera was used for capturing the videos with resolution 640x480. The speed of the videos was 30 frames per second. A few more videos were collected and kept aside to test the model.

*Data Preparation:* The data collected were raw video feeds and they needed to be processed to create the desired training data. To create the training data, the videos were sent through the MediaPipe library. The pre-trained model in this library is capable of extraction 33 key points of human body when it is fed with an image of a human pose. These 33 points were basically the coordinates of the major key points of the human body identified by the pre-trained deep learning model. The pose landmarks are shown in Figure 1.

The model's inference was quite fast to extract the key points from every frame of the video. The output of the model consisted of X, Y and Z co-ordinate of the key points and for this study, the Z co-ordinate was not taken into the consideration. This was done mainly to reduce the total number of co-ordinates per frame (66 instead of 99 and thus reducing the curse of dimensionality) as the study focused on the road crossing behaviour of pedestrians who were using the shortest route to cross the road, i.e. perpendicular to the edge of the road.

In such a situation, Z axis would not have had any significant impact. After the coordinates were extracted (total 66 in number per frame), those were stored in a CSV file. Moreover, the frames were also looked at carefully to understand the time at which the pedestrian started crossing the road. This was done manually by observing the frames individually. The videos were of length roughly 10 seconds each. Hence, for each video around 300 frames were available.

For each frame it was required to understand whether the person was about to cross the road or not. If a road crossing tendency was there in any frame, the fame was labelled with 1 and if the tendency was absent, it was labelled with 0. This entire labelling was done manually. The entire process in shown in Figure 2. In this way 60 individual CSV files were created with 66 predictor variables and one binary target variables

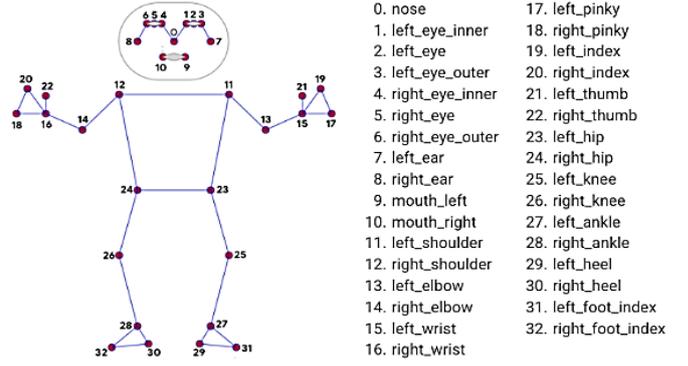

**Figure 2:** Key points detected by the MediaPipe library

*Sequence Modelling:* Analysing pedestrian intent to cross the road is a task which can be solved by employing different sequence modelling techniques. Prior to the advent of deep learning methods, the popular sequence modelling methods were Hidden Markov Model (HMM) [18] and Conditional Random field (CRF) [19]. But these models are more closely associated with statistical models whereas deep learning models are more equipped to deal with nonlinear patterns associated with the data. Particularly, Long Short Tern Memory (LSTM) and Gated Recurrent Unit (GRU) models are well equipped to map nonlinear patterns with much longer sequences. LSTM efficiently deals with the longer sequence of data by adopting 4 gates within the cell. Each gate processes the information differently leading to reduction of the vanishing (or exploding) gradient problem associated with simple Recurrent Neural Network (RNN). The 4 gates are

$$Forget\ gate: f_t = Sigmoid([h_{t-1}, x_t]w_f + b_f)$$

$$input\ gate: i_t = Sigmoid([h_{t-1}, x_t]w_i + b_i)$$

$$\tilde{C}_t = tanh([h_{t-1}, x_t]w_C + b_C)$$

$$update\ gate: C_t = f_t C_{t-1} + i_t \tilde{C}_t$$

$$output\ gate: o_t = Sigmoid([h_{t-1}, x_t]w_o + b_o)$$

$$h_t = o_t \tanh(C_t)$$

where $h_t, C_t, x_t$ are the hidden state, cell state and current input (at time t) respectively. $w_f, w_i, w_o$ are the respective weights associated with the gates and $b_f, b_i, b_o$ are biases associated with the individual gates respectively. $\tilde{C}_t$ is the candidate set determining the extent of current input to be contained in the current cell state. Its value ranges from -1 to +1. Details of the gates are available in the original paper of LSTM [20].

The other model under consideration was Gated Recurrent Unit (GRU). Unlike LSTM, GRU has 2 gates making it faster than LSTM during inference. Interestingly, performance of GRU is often found to be comparable to LSTM and there are instances where GRU performed better than LSTM. The details of the gates are given below.

$$reset\ gate: r_t = Sigmoid([h_{t-1}, x_t]w_r + b_r)$$

$$update\ gate: z_t = Sigmoid([h_{t-1}, x_t]w_z + b_z)$$

$$\tilde{h}_t = tanh([r_t \odot h_{t-1}, x_t]w_h + b_h)$$

$$h_t = (1 - z_t) \odot \tilde{h}_t + z_t \odot h_{t-1}$$

Here, $r_t$ is the output after the reset operation which determines how much of the past information to be forgotten when dealing with current data $x_t$, $z_t$ is the update gate which determines the amount of pervious input required to be kept and by how much the same needs to be updated with the current data $x_t$. $\tilde{h}_t$ and $h_t$ are the candidate hidden state and final hidden state respectively. $\odot$ represents the elementwise multiplication (Hadamard product).

The third model used in this study was one-dimensional convolutional network. One dimensional convolutional network works by applying convolution operation on a 1D data over an input sequence. The generated feature map is afterward mapped with the target variable for training the model. Subsequently, the trained model is used for inferences based on test dataset. Mathematically, the 1D convolutional network works as per the given equation below.

$$y_t = \sum_{i=0}^{k-1} w_i x_{t-i} + b$$

where $x_t$ is the input at time stamp $t$, $w_t$ is the weight, $k$ is the kernel size, $b$ is the bias and $y_t$ is the output at time stamp $t$. 1D convolutional network is much faster than LSTM and GRU and hence it is also used in places where speed of inference does play an important role. The system used for this study was a laptop with Windows 11 operating system coupled with an i5 CPU (11th Gen) with 3.20GHz-3.19 GHz speed. The laptop had NVIDIA GTX 1650 GPU (4 GB dedicated memory) and 16 GB RAM. The deep learning framework used was PyTorch (v 2.5.1) with CUDA version 11.8.

## IV. ANALYSIS

A video is essentially a series of still images, or frames, and thus, analysing a video can be viewed as a sequential analysis of these frames. When poses within these frames are analysed in sequence, it becomes possible to predict whether a person is likely to engage in a particular action. This necessitates the use of a model capable of sequential data analysis, and for this study, Long Short-Term Memory (LSTM), Gated Recurrent Unit (GRU) and 1D Convolutional Neural Network (1D CNN) models were employed.

In the present study, a sequence length of 15 frames was selected. The task was to classify the 16th frame based on the preceding 15 frames, specifically determining whether a pedestrian was crossing the road or not. Some videos included pedestrians who initially appeared to be crossing but then decided to backtrack. In such cases, the frames depicting the backtracking behaviour were reclassified from 'crossing' to 'not crossing.' This adjustment allowed the models to learn the tendencies of pedestrians to change their decision while crossing the road.

The LSTM and GRU models were trained with similar configurations. The architecture for both included either a single or stacked LSTM or GRU layer, followed by a dropout layer and an output layer with a sigmoid activation function to

```
==========================================================================
Layer (type:depth-idx)                   Output Shape              Param #
==========================================================================
LSTMModel                                [32, 1]                   --
├─LSTM: 1-1                              [32, 15, 50]              44,000
├─Linear: 1-2                            [32, 1]                   51
├─Sigmoid: 1-3                           [32, 1]                   --
==========================================================================
Total params: 44,051
Trainable params: 44,051
Non-trainable params: 0
Total mult-adds (M): 21.12
==========================================================================
Input size (MB): 0.13
Forward/backward pass size (MB): 0.19
Params size (MB): 0.18
Estimated Total Size (MB): 0.50
==========================================================================
```

**Figure 3**: Architecture of LSTM model

```
==========================================================================
Layer (type:depth-idx)                   Output Shape              Param #
==========================================================================
GRUModel                                 [32, 1]                   --
├─GRU: 1-1                               [32, 15, 50]              33,000
├─Linear: 1-2                            [32, 1]                   51
├─Sigmoid: 1-3                           [32, 1]                   --
==========================================================================
Total params: 33,051
Trainable params: 33,051
Non-trainable params: 0
Total mult-adds (M): 15.84
==========================================================================
Input size (MB): 0.13
Forward/backward pass size (MB): 0.19
Params size (MB): 0.13
Estimated Total Size (MB): 0.45
==========================================================================
```

**Figure 4**: Architecture of GRU model

```
==========================================================================
Layer (type:depth-idx)                   Output Shape              Param #
==========================================================================
CNNModel                                 [32, 1]                   --
├─Conv1d: 1-1                            [32, 50, 13]              9,950
├─ReLU: 1-2                              [32, 50, 13]              --
├─Linear: 1-3                            [32, 1]                   51
├─Sigmoid: 1-4                           [32, 1]                   --
==========================================================================
Total params: 10,001
Trainable params: 10,001
Non-trainable params: 0
Total mult-adds (M): 4.14
==========================================================================
Input size (MB): 0.13
Forward/backward pass size (MB): 0.17
Params size (MB): 0.04
Estimated Total Size (MB): 0.33
==========================================================================
```

**Figure 5**: Architecture of the CNN model

evaluate the binary cross-entropy loss. The intermediate layers used ReLU activation. Through manual hyperparameter tuning, it was found that both models performed best with a stack of two LSTM or GRU layer consisting of 50 hidden units. Additionally, the dropout layer yielded optimal results with a dropout probability of 0.5.

The sequential analysis approach utilized in this study relied on the strengths of LSTM and GRU networks. These models are particularly well-suited for handling sequences due to their ability to maintain and update a memory of previous frames, which is crucial for understanding temporal dependencies in video data. The single-layer architecture with 50 hidden units was chosen after experimentation, ensuring a balance between model complexity and performance. The dropout layer, with a probability of 0.5, served to minimize overfitting by randomly omitting certain neurons during training, thereby enhancing the model's generalizability to unseen data

By training on sequences of 15 frames and predicting the class of the 16th frame, the models effectively learned to anticipate pedestrian behaviour. The reclassification of backtracking frames was an important step, as it ensured that the models were not just learning the straightforward task of crossing but also the more nuanced behaviour of changing one's mind mid-action. This capability is critical in real-world scenarios, such as autonomous driving, where understanding pedestrian intentions accurately can significantly enhance safety and decision-making processes. The architectures are given in Figure 3 and Figure 4. Figure 3 depicts the architecture of the LSTM model with a single LSTM layer with 50 hidden nodes and Figure 4 depicts the GRU model having the same 50 hidden nodes. Since LSTM has 4 gates, the number of parameters of the LSTM layer is more than that of the GRU layer as GRU works with only 2 gates.

The CNN model used in this study used a kernel size of 3 in a single convolution layer followed by global average pooling operation and the dense layer. The architecture of the 1D CNN model is shown in Figure 5. It can be seen quite easily that the number of parameters of the CNN model is lowest among the three models.

All the models were trained with 10 epochs and the training and validation losses were compared. The same thing was also done with the training and validation accuracies. The movement of losses and the accuracies are shown in Figure 6 and Figure 7. Figure 8 shows how the Area Under the ROC Curve (AUC) scores varied with respect to the epochs. Prior to training the model, the dataset was split into two parts as trainset and testset with 90:10 ratio. The number of datapoints for the trainset was 15585 and for the test dataset it number was 1823. During training, the trainset was further split into train and validation sets with 80:20 ratio. Thus, for testing the three models, a common test set was kept ready. During training, the model with highest AUC score was stored for all the three types of models. AUC was considered along with accuracy because of the presence of class imbalance within the dataset. Figure 6 shows that the training loss went down gradually with the epoch for all the three models and the same trend was there with the validation process. The GRU model reached the lowest validation loss. The CNN model, with only one convolution layer and global pooling process showed significantly good result even though its performance stayed below than the LSTM and GRU model.

Figure 7 shows how the accuracies of the three models fluctuated with respect to the epochs. The accuracies of all the models got more stabilized with epochs. The interesting part observed in this plot is that the training and validation accuracies are not very different from each other at every epoch suggesting very little scope of overfitting. It is to be noted that the dataset contained class imbalance and hence, for better judgement of the models, the AUC scores were required to be checked because AUC is robust to the presence of class imbalances. Figure 8 shows that the AUC scores got saturated gradually to around 0.73 for CNN model whereas for LSTM and GRU the AUC scores got almost saturated at around 0.82. In this plot also it is seen clearly that training and validation AUC scores are quite matching with each other, suggesting a good fit for the models with no significant overfitting. What is interesting in all these three plots is that the CNN model with much lesser capacity (in terms of number of parameters) followed the LSTM and GRU models' performances closely.

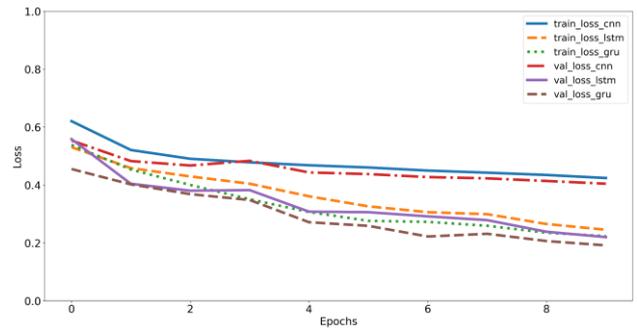

**Figure 6**: Movement of training and validation losses of the three models with respect to epochs

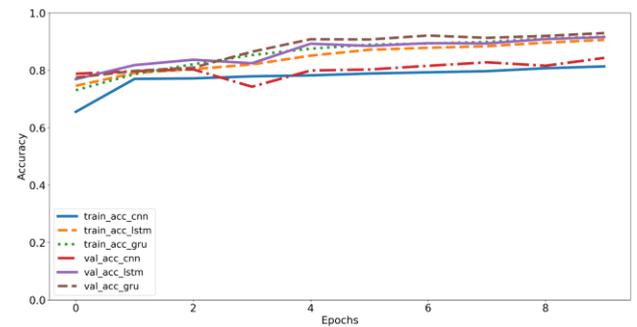

**Figure 7**: Movement of training and validation accuracies of the three models with respect to epochs

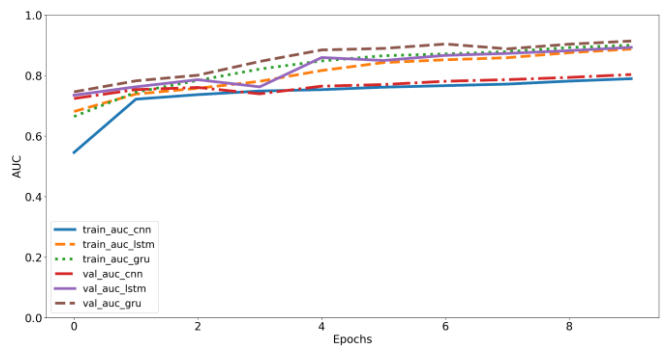

**Figure 8**: Movement of training and validation AUC of the three models with respect to epochs

The choice of hyperparameters played a very crucial role in the performance of all the models. The hidden unit size of 50 was found to be optimal, providing enough capacity to learn complex patterns for both the LSTM and the GRU model. The CNN model contained only one CNN layers with global average pooling operation. For the CNN models, average pooling smoothened out the irregularities leading to stable learning without any sign of model overfitting. The shallow architecture prevented the chance of covariate shift and thus making the model robust to high degree of overfitting.

As mentioned earlier, the best model was saved during the training process, and the best models were applied on the test dataset. The Table 1 below shows the performance of the models on the test dataset. The test dataset had 1823 number of data points and 1D CNN was faster than both LSTM and GRU model during inference. The accuracy was less than that of LSTM and GRU but for real time analysis, CNN based model would be more preferred even with slightly lesser accuracies.

TABLE 1: PERFORMANCE OF THE MODELS ON THE TEST DATASET

| Model | Test Accuracy | Test AUC | Inference time |
|-------|---------------|----------|----------------|
| 1D CNN | 81.95% | 74.27% | 1 ms |
| LSTM | 87.00% | 86.38% | 3 ms |
| GRU | 85.52% | 89.24% | 2 ms |

On the same laptop, when the models were run using CPU only, for a roughly 12 second video the models were taking less than 20 seconds to produce a new video with crossing probabilities based on the pretrained Mediapipe model and the custom sequence models trained with LSTM, GRU and 1D CNN. This outcome was found to be much faster than some other previous research where the average inference time per frame was 100 milliseconds or more. The end-to-end Mediapipe with sequence model framework was able to process frames at an average speed of 43 milliseconds per frame. This included the inference time of Mediapipe model and other intermediate data handling processes for the sequence model. The code was not optimized for speed, but it could produce results with some lags.

## V. CONCLUSION

The study demonstrates the feasibility and effectiveness of using LSTM and GRU and 1D CNN models for predicting pedestrian behaviour from video sequences. The study demonstrated that all the models could predict the intent of road crossing based on the output of MediaPipe. As per the analysis, it came out that GRU was most accurate in making prediction and faster than LSTM but was slower than 1D CNN model. An average inference time per frame was estimated around 43 milliseconds and it was not fast enough for a 30 FPS video. However, for a lesser FPS video, the framework could work in real time. The ability to accurately anticipate whether a pedestrian will cross the road or change their mind is invaluable for enhancing the safety and decision-making capabilities of autonomous driving systems. The insights gained from this study can inform future research and development in the field of video-based behaviour prediction, contributing to the advancement of intelligent transportation systems. The same method can also be used in many other situations where sequential analysis could provide valuable insights based on human movements. The same process can also be used in sport analytics or even detecting suspicious movements near restricted areas. The study was not out of limitations. One of the key limitations of the study was to accommodate multiple people to predict their road crossing intents. For rather simplicity, the videos were captured with only one pedestrian. However, in real life scenario, multiple pedestrians would be there in a frame and all of them are required to be analysed. The second limitation was to achieve a lesser inference time with these models. In future research, these limitations would be tried to be addressed for accurate prediction at higher speed.